\documentclass[a4paper,runningheads]{llncs}

\usepackage{alltt}
\usepackage{caption}
\usepackage{paralist}
\usepackage{nicefrac}
\usepackage{booktabs}
\usepackage{algorithm}
\usepackage{algorithmic}
\usepackage{xspace}
\usepackage{amsmath}
\usepackage{marvosym}
\usepackage{wasysym}
\usepackage{verbatim}
\usepackage{subfigure}
\usepackage{hyperref}
\usepackage{multirow}
\usepackage{cite}
\usepackage[capitalize]{cleveref}
\usepackage[per-mode=symbol,detect-all]{siunitx}
\usepackage{graphics}
\usepackage{amssymb}
\usepackage{wrapfig}
\usepackage{enumitem}
\usepackage{xcolor}
\usepackage{booktabs}

\definecolor{navy}{RGB}{0,0,128}

\newcommand{\relu}{\text{ReLU}\xspace{}}
\newcommand{\sign}{\text{sign}\xspace{}}

\usepackage{tikz,ifthen,pgfplots}
\usetikzlibrary{arrows,trees,backgrounds,automata,shapes,decorations,plotmarks,fit,calc,positioning,shadows,chains}
\tikzstyle{every pin edge}=[<-,shorten <=1pt]
\tikzstyle{neuron}=[circle,fill=black!25,minimum size=17pt,inner sep=0pt]
\tikzstyle{input neuron}=[neuron, fill=green!50]
\tikzstyle{output neuron}=[neuron, fill=red!50]
\tikzstyle{hidden neuron}=[neuron, fill=blue!50]
\tikzstyle{merged neuron}=[neuron, fill=orange!50]
\tikzstyle{annot} = [text width=6em, text centered]
\tikzstyle{nnedge} = [-{stealth},shorten >=0.1cm, shorten <=0.05cm,line width=0.8pt,black]
\usetikzlibrary{calc}

\newcommand{\allvars}{\mathcal{X}}
\newcommand{\ub}{u}
\newcommand{\lb}{l}
\newcommand{\signSet}{S}
\newcommand{\reluSet}{R}
\newcommand{\assignment}{\alpha{}}

\newcommand{\sat}{\texttt{SAT}}
\newcommand{\unsat}{\texttt{UNSAT}}
\newcommand{\PComplexity}{\texttt{P}}
\newcommand{\NPComplexity}{\texttt{NP}}
\newcommand{\PSPACEComplexity}{\texttt{PSPACE}}

\newcommand{\snc}{S\textup{\&}C\xspace{}}

\newcommand{\drule}[2]{
\renewcommand{\arraystretch}{1.2}
\(\begin{array}{c}
#1 \\
\hline 
#2
\end{array}\)
}

\newcommand{\rulename}[1]{\ensuremath{\mathsf{#1}}\xspace}
\newcommand{\irulename}[2]{\ensuremath{\mathsf{#1}_{#2}}\xspace}

\newcommand{\reluCorrectB}{\irulename{ReluCorrect}{b}}
\newcommand{\reluCorrectF}{\irulename{ReluCorrect}{f}}
\newcommand{\signCorrectN}{\irulename{SignCorrect}{-}}
\newcommand{\signCorrectP}{\irulename{SignCorrect}{+}}

\newcommand{\success}{\rulename{Success}}

\newcommand{\signSplit}{\rulename{SignSplit}}

\newcommand{\reluSplit}{\rulename{ReluSplit}}

\newcommand{\updateOperation}{\textit{update}}
\newcommand{\addEqOperation}{\textit{addEq}}

\usepackage{marvosym}

\usepackage[firstpage]{draftwatermark}
\SetWatermarkText{\hspace*{6.5in}\raisebox{6.5in}{\includegraphics[scale=0.1]{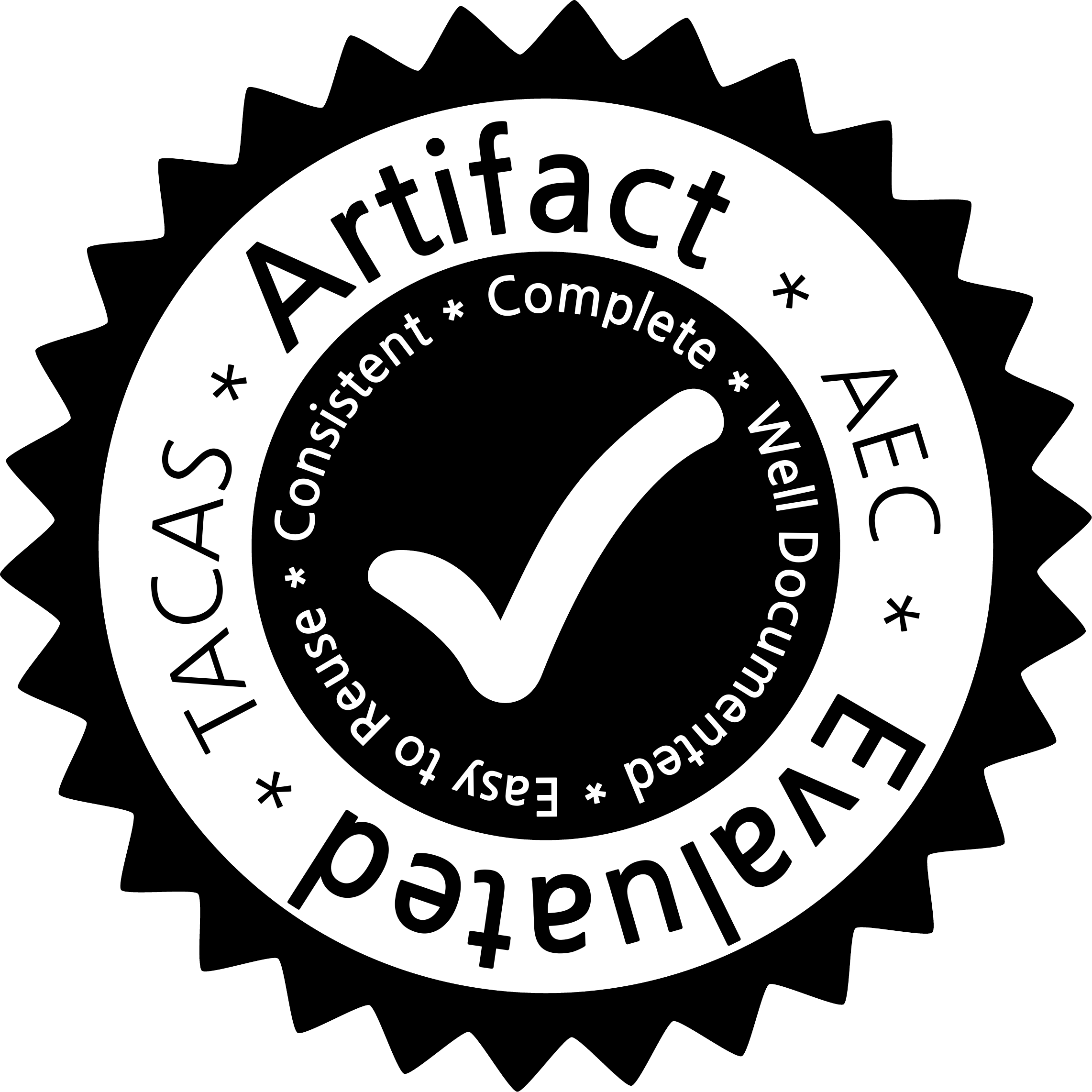}}}
\SetWatermarkAngle{0}

\authorrunning{G. Amir et al.}

\begin{document}

\title{An SMT-Based Approach for Verifying Binarized Neural Networks}

\author{
  Guy Amir\inst{1} \and
  Haoze Wu\inst{2} \and
  Clark Barrett\inst{2} \and
  Guy Katz\inst{1}[\Letter]
}
\institute{
  The Hebrew University of Jerusalem, Jerusalem, Israel \\
  \email{ \{guy.amir2, g.katz\}@mail.huji.ac.il}\\
  \and
  Stanford University, Stanford, USA \\
  \email{ \{haozewu, barrett\}@cs.stanford.edu }\\
}

\maketitle

\begin{abstract}
  Deep learning has emerged as an effective approach for creating
  modern software systems, with neural networks often surpassing
  hand-crafted systems. Unfortunately, neural networks are known
  to suffer from various safety and security issues. Formal
  verification is a promising avenue for tackling this difficulty, by
  formally certifying that networks are correct. We propose an
  SMT-based technique for verifying \emph{binarized neural networks}
  --- a popular kind of neural network, where some weights have been
  binarized in order to render the neural network more memory and
  energy efficient, and quicker to evaluate. One novelty of our
  technique is that it allows the verification of neural networks that
  include both binarized and non-binarized components.  Neural network
  verification is computationally very difficult, and so we propose
  here various optimizations, integrated into our SMT procedure as
  deduction steps, as well as an approach for parallelizing
  verification queries. We implement our technique as an extension to
  the Marabou framework, and use it to evaluate the approach on popular
  binarized neural network architectures.
\end{abstract}

\section{Introduction}
In recent years, \emph{deep neural networks}
(\emph{DNNs})~\cite{goodfellow2016deep} have revolutionized the state
of the art in a variety of tasks, such as image
recognition~\cite{krizhevsky2012imagenet, CiMeSc12}, text
classification~\cite{lai2015recurrent}, and many others. These DNNs,
which are artifacts that are generated automatically from a set of
training data, generalize very well --- i.e., are very successful at
handling inputs they had not encountered previously. The success of
DNNs is so significant that they are increasingly being incorporated
into highly-critical systems, such as autonomous vehicles and
aircraft~\cite{BoDeDwFiFlGoJaMoMuZhZhZhZi16,JuLoBrOwKo16}.

In order to tackle increasingly complex tasks, the size of modern DNNs
has also been increasing, sometimes reaching many millions of
neurons~\cite{SiZi15}. Consequently, in some domains, DNN size has become a
restricting factor: huge networks have a large memory footprint, and
evaluating them consumes both time and energy. Thus, resource-efficient
networks are required in order to allow DNNs to be deployed on
resource-limited, embedded devices~\cite{molchanov2016pruning,HaMaDa16}.

One promising approach for mitigating this problem is via DNN
\emph{quantization}~\cite{hubara2017quantized, BaStKoQnn20}. Ordinarily, each edge
in a DNN has an associated weight, typically stored as a 32-bit
floating point number. In a quantized network, these weights are
stored using fewer bits. Additionally, the \emph{activation functions}
used by the network are also quantized, so that their outputs consist
of fewer bits.  The network's memory footprint thus becomes
significantly smaller, and its evaluation much quicker and cheaper.
When the weights and activation function outputs are represented using
just a single bit, the resulting network is called a \emph{binarized
  neural network} (\emph{BNN})~\cite{hubara2016binarized}. BNNs are a
highly popular variant of a quantized
DNN~\cite{LiTaAn2016,ZhMoChFr2017,YaShXiTiLiDeHuHu2019,ChZhZhZhLiJiDoGu20},
as their computing time can be up to 58 times faster, and their memory
footprint 32 times smaller, than that of traditional
DNNs~\cite{rastegari2016xnor}. There are also network architectures in
which some parts of the network are quantized, and others are
not~\cite{rastegari2016xnor}. While quantization leads to some loss of
network precision, quantized networks are sufficiently precise in many
cases~\cite{rastegari2016xnor}.

In recent years, various security and safety issues have been observed
in DNNs~\cite{SzZaSuBrErGoFe13,KaBaDiJuKo21ReluplexRevised}. This has
led to the development of a large variety of verification tools and
approaches
(e.g.,~\cite{KaBaDiJuKo21ReluplexRevised,HuKwWaWu17,GeMiDrTsChVe18,WaPeWhYaJa18},
and many others). However, most of these approaches have not focused
on binarized neural networks, although they are just as vulnerable to
safety and security concerns as other DNNs. Recent work has shown that
verifying quantized neural networks is
\PSPACEComplexity{}-hard~\cite{HeKeZiScalable20}, and that it requires
different methods than the ones used for verifying non-quantized
DNNs~\cite{GiMiHeThMa20}. The few existing approaches that do handle
binarized networks focus on the \emph{strictly binarized} case, i.e.,
on networks where \emph{all} components are binary, and verify them
using a SAT solver
encoding~\cite{narodytska2017verifying,jia2020efficient}. Neural
networks that are only partially binarized~\cite{rastegari2016xnor}
cannot be readily encoded as SAT formulas, and thus verifying these
networks remains an open problem.

Here, we propose an SMT-based~\cite{BaTi18Smt} approach and tool for
the formal verification of binarized neural networks. We build on top
of the Reluplex
algorithm~\cite{KaBaDiJuKo21ReluplexRevised},\footnote{\cite{KaBaDiJuKo21ReluplexRevised}
  is a recent extended version of the original Reluplex
  paper~\cite{KaBaDiJuKo17Reluplex}.}  and extend it so that it can
support the \emph{sign} function,
\[
  \sign(x) =
  \begin{cases}
    x < 0 & -1 \\
    x \geq 0 & 1.
    \end{cases}
\]
We show how this extension, when integrated into Reluplex, is
sufficient for verifying BNNs.  To the best of our knowledge, the
approach presented here is the first capable of verifying BNNs that
are not strictly binarized.  Our technique is implemented as an
extension to the open-source Marabou
framework~\cite{KaHuIbJuLaLiShThWuZeDiKoBa19Marabou, MarabouGitRepo}.  We discuss the
principles of our approach and the key components of our implementation. We
evaluate it both on the XNOR-Net BNN
architecture~\cite{rastegari2016xnor}, which combines binarized and
non-binarized parts, and on a strictly binarized network.

The rest of this paper is organized as follows. In
Section~\ref{sec:background}, we provide the necessary background on
DNNs, BNNs, and the SMT-based formal verification of DNNs. Next, we
present our SMT-based approach for supporting the sign activation
function in Section~\ref{sec:supportingSigns}, followed by details on
enhancements and optimizations for the approach in
Section~\ref{sec:optimizations}. We discuss the implementation of our
tool in Section~\ref{sec:tool}, and its evaluation in
Section~\ref{sec:evaluation}. Related work is discussed in
Section~\ref{sec:relatedWork}, and we conclude in
Section~\ref{sec:conlcusion}.

\section{Background}
\label{sec:background}  

\subsubsection{Deep Neural Networks.}
A deep neural network (DNN) is a directed graph, where the nodes (also
called neurons) are organized in layers. The first layer is
the \emph{input layer}, the last layer is the \emph{output layer}, and
the intermediate layers are the \emph{hidden layers}. When the network
is evaluated, the input neurons are assigned initial values
(e.g., the pixels of an image), and these values
are then propagated through the network, layer by layer, all the way
to the output layer. The values of the output neurons determine the
result returned to the user: often, the neuron with the greatest value
corresponds to the output class that is returned. A network is called
\textit{feed-forward} if outgoing edges from neurons in layer $i$ can only lead
to neurons in layer $j$ if $j>i$. For simplicity, we will assume here
that outgoing edges from layer $i$ only lead to the consecutive layer,
$i+1$.

Each layer in the neural network has a \emph{layer type}, which
determines how the values of its neurons are computed (using the
values of the preceding layer's neurons). One common type is the
\emph{weighted sum} layer: neurons in this layer are computed as a
linear combination of the values of neurons from the preceding layer, according
to predetermined edge weights and biases.  Another common type of layer is
the \emph{rectified linear unit} (\emph{ReLU}) layer, where each node
$y$ is connected to precisely one node $x$ from the preceding layer,
and its value is computed by $y=\relu{}(x)=\max(0,x)$. The \emph{max-pooling} layer 
is also common: each neuron $y$ in this layer is connected to
multiple neurons $x_1,\ldots,x_k$ from the preceding layer, and its value
is given by $y=\max(x_1,\ldots,x_k)$.

More formally, a DNN $N$ with $k$ inputs and $m$ outputs is a mapping
$\mathbb{R}^k\rightarrow\mathbb{R}^m$. It is given as a sequence of
layers $L_1,\ldots, L_n$, where $L_1$ and $L_n$ are the input and
output layers, respectively. We denote the size of layer $L_i$ as
$s_i$, and its individual neurons as $v_i^1,\ldots,v_i^{s_i}$. We use
$V_i$ to denote the column vector $[v_i^1,\ldots,v_i^{s_i}]^T$.
During evaluation, the input values $V_1$ are given, and
$V_2,\ldots,V_n$ are computed iteratively. The network also includes a
mapping $T_N:\mathbb{N}\rightarrow\mathcal{T}$, such that $T(i)$
indicates the \emph{type} of hidden layer $i$. For our purposes, we focus 
on layer types $\mathcal{T}=\{\text{weighted sum}, \relu{}, \max\}$, but
of course other types could be included. If
$T_n(i)=\text{weighted sum}$, then layer $L_i$ has a weight matrix
$W_i$ of dimensions $s_{i}\times s_{i-1}$ and a bias vector $B_i$ of
size $s_i$, and its values are computed as $V_i=W_i\cdot
V_{i-1}+B_i$. For $T_n(i)=\relu{}$, the \relu{} function is applied to
each neuron, i.e. $v_i^j=\relu{}(v_{i-1}^j)$ (we required that
$s_i = s_{i-1}$ in this case). If $T_n(i)=\max$, then each neuron
$v_i^j$ in layer $L_i$ has a list $src$ of source indices, and its
value is computed as $v_i^j=\max_{k\in src}v_{i-1}^k$.

\begin{wrapfigure}[8]{r}{6.5cm}
  \vspace{-1.2cm}
  \begin{center}
    \scalebox{0.75} {
      \def\layersep{2.0cm}
    \begin{tikzpicture}[shorten >=1pt,->,draw=black!50, node distance=\layersep,font=\footnotesize]

      \node[input neuron] (I-1) at (0,-1) {$v^1_1$};
      \node[input neuron] (I-2) at (0,-2.5) {$v^2_1$};

      \node[hidden neuron] (H-1) at (\layersep,-1) {$v^1_2$};
      \node[hidden neuron] (H-2) at (\layersep,-2.5) {$v^2_2$};

      \node[hidden neuron] (H-3) at (2*\layersep,-1) {$v^1_3$};
      \node[hidden neuron] (H-4) at (2*\layersep,-2.5) {$v^2_3$};

      \node[output neuron] at (3*\layersep, -1.75) (O-1) {$v^1_4$};

      \draw[nnedge] (I-1) --node[above] {$1$} (H-1);
      \draw[nnedge] (I-1) --node[above, pos=0.3] {$\ -5$} (H-2);
      \draw[nnedge] (I-2) --node[below, pos=0.3] {$2$} (H-1);
      \draw[nnedge] (I-2) --node[below] {$1$} (H-2);

      \draw[nnedge] (H-1) --node[above] {$\relu$} (H-3);
      \draw[nnedge] (H-2) --node[below] {$\relu$} (H-4);

      \draw[nnedge] (H-3) --node[above] {$1$} (O-1);
      \draw[nnedge] (H-4) --node[below] {$-2$} (O-1);

      \node[below=0.05cm of H-1] (b1) {$+1$};
      \node[below=0.05cm of H-2] (b2) {$+2$};

      \node[annot,above of=H-1, node distance=0.8cm] (hl1) {Weighted sum};
      \node[annot,above of=H-3, node distance=0.8cm] (hl2) {ReLU };
      \node[annot,left of=hl1] {Input };
      \node[annot,right of=hl2] {Output };
    \end{tikzpicture}
    }
    \captionsetup{size=small}
    \captionof{figure}{A toy DNN.}
    \label{fig:toyDnn}
  \end{center}
\end{wrapfigure}
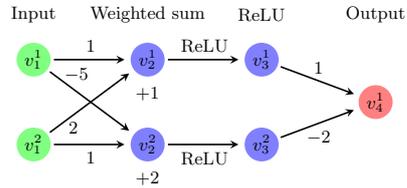

A simple illustration appears
in Fig.~\ref{fig:toyDnn}. This network has a weighted sum layer and a
\relu{} layer as its hidden layers, and a weighted sum layer as
its output layer. For the weighted sum layers, the
weights and biases are listed in the figure. On input $V_1=[1, 2]^T$,
the first layer's neurons evaluate to $V_2=[6,-1]^T$. After \relu{}s
are applied, we get $V_3=[6,0]^T$, and finally the output is $V_4=[6]$.

\subsubsection{Binarized Neural Networks.}

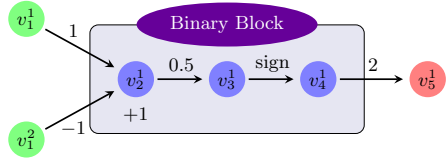
\begin{wrapfigure}[9]{r}{7.5cm}
  \vspace{-1.4cm}
  \begin{center}
    \scalebox{0.8} {
      \def\layersep{1.5cm}
    \begin{tikzpicture}[shorten >=1pt,->,draw=black!50, node distance=\layersep,font=\footnotesize]

      \node[input neuron] (I-1) at (0,-1) {$v^1_1$};
      \node[input neuron] (I-2) at (0,-3) {$v^2_1$};

      \node[hidden neuron] (H-1) at (\layersep + 0.3cm,-2) {$v^1_2$};

      \node[hidden neuron] (H-2) at (2*\layersep+ 0.3cm,-2) {$v^1_3$};
      \node[hidden neuron] (H-3) at (3*\layersep+ 0.3cm,-2) {$v^1_4$};

      \node[output neuron] at (4*\layersep+ 0.6cm, -2) (O-1) {$v^1_5$};

      \draw[nnedge] (I-1) --node[above,pos=0.4] {$1$} (H-1);
      \draw[nnedge] (I-2) --node[below,pos=0.4] {$-1$} (H-1);

      \draw[nnedge] (H-1) --node[above] {$0.5$} (H-2);
      \draw[nnedge] (H-2) --node[above] {$\sign$} (H-3);

      \draw[nnedge] (H-3) --node[above] {$2$} (O-1);

      \node[below=0.05cm of H-1] (b1) {$+1$};

      \begin{pgfonlayer}{background}

        \tikzstyle{background_rectangle}=[rounded corners, fill = navy!10]

        \draw[background_rectangle]
        ($(H-1.west) + (-0.3*\layersep, -0.9cm ) $)
        rectangle
        ($(H-3.east) + ( 0.3*\layersep, 0.9cm ) $);

        \tikzstyle{fancytitle} =[fill=blue!60!red, text=white, ellipse]
        
        \node[fancytitle] at ($(H-2.north) +
        (0,0.6cm)$) {Binary Block};

      \end{pgfonlayer}

    \end{tikzpicture}
    }
    \captionsetup{size=small}
    \captionof{figure}{A toy BNN with a single binary block composed
    of three layers: a weighted sum layer, a batch normalization
      layer, and a \sign{} layer.}
    \label{fig:toyBnn}
  \end{center}
\end{wrapfigure}

In a \emph{binarized neural network} (\emph{BNN}), the layers are
typically organized into binary \emph{blocks}, regarded as units with
binary inputs and outputs. Following the definitions of Hubara et
al.~\cite{hubara2016binarized} and Narodytska et
al.~\cite{narodytska2017verifying}, a binary block is comprised of three layers:
\begin{inparaenum}[(i)] 
\item a \emph{weighted sum} layer, where each entry of the weight matrix $W$
  is either $1$ or $-1$;
\item a \emph{batch normalization} layer, which normalizes the values from
  its preceding layer (this layer can be regarded as a weighted sum
  layer, where the weight matrix $W$ has real-valued entries in its
  diagonal, and 0 for all other entries); and
\item a \emph{sign} layer, which applies the \sign{} function to each
  neuron in the preceding layer.
\end{inparaenum}
Because each block ends with a sign layer, its output is always a
binary vector, i.e. a vector whose entries are $\pm 1$. Thus, when
several binary blocks are concatenated, the inputs and outputs of each
block are always binary. Here, we call a network \emph{strictly
  binarized} if it is composed solely of binary blocks (except for the
output layer).  If the network contains binary blocks but also
additional layers (e.g., \relu{} layers), we say that it is a
\emph{partially binarized} neural network. BNNs can be made to fit into our
definitions by extending the set $\mathcal{T}$ to include the \sign{}
function. An example appears in Fig.~\ref{fig:toyBnn}; for input
$V_1=[-1,3]^T$, the network's output is $V_5=[-2]$.

\subsubsection{SMT-Based Verification of Deep Neural Networks.}

Given a DNN $N$ that transforms an input vector $x$ into an output
vector $y=N(x)$, a pre-condition $P$ on $x$, and a post-condition $Q$
on $y$, the \emph{DNN verification
  problem}~\cite{KaBaDiJuKo21ReluplexRevised} is to determine whether there
exists a concrete input $x_0$ such that $P(x_0)\wedge
Q(N(x_0))$. Typically, $Q$ represents an undesirable output of the
DNN, and so the existence of such an $x_0$ constitutes a
counterexample. A sound and complete verification engine should return a suitable $x_0$
if the problem is satisfiable (\sat), or reply that it is
unsatisfiable (\unsat). As in most DNN verification literature, we
will restrict ourselves to the case where $P$ and $Q$ are conjunctions
of linear constraints over the input and output neurons,
respectively~\cite{KaBaDiJuKo21ReluplexRevised,GeMiDrTsChVe18,WaPeWhYaJa18}.

Here, we focus on an SMT-based approach for DNN verification, which
was introduced in the Reluplex algorithm~\cite{KaBaDiJuKo21ReluplexRevised}
and extended in the Marabou
framework~\cite{KaHuIbJuLaLiShThWuZeDiKoBa19Marabou, MarabouGitRepo}. It entails
regarding the DNN's node values as variables, and the verification
query as a set of constraints on these variables. The solver's goal is
to find an assignment of the DNN's nodes that satisfies $P$ and
$Q$. The constraints are partitioned into two sets: \emph{linear
  constraints}, i.e. equations and variable lower and upper bounds,
which include the input constraints in $P$, the output constraints in
$Q$, and the weighted sum layers within the network; and
\emph{piecewise-linear constraints}, which include the activation
function constraints, such as \relu{} or max constraints. The linear
constraints are easier to solve (specifically, they can be phrased as
a linear program~\cite{BaIoLaVyNoCr16}, solvable in polynomial time);
whereas the piecewise-linear constraints are more difficult, and
render the problem \NPComplexity{}-complete~\cite{KaBaDiJuKo21ReluplexRevised}. We observe
that \sign{} constraints are also piecewise-linear.

In Reluplex, the linear constraints are solved iteratively, using a
variant of the Simplex algorithm~\cite{Dantzig1963}. Specifically,
Reluplex maintains a variable assignment, and iteratively corrects the
assignments of variables that violate a linear constraint. Once the
linear constraints are satisfied, Reluplex attempts to correct any
violated piecewise-linear constraints --- again by making iterative 
adjustments to the assignment. If these steps re-introduce
violations in the linear constraints, these constraints
are addressed again. Often, this process converges; but if it
does not, Reluplex performs a \emph{case split}, which transforms one
piecewise-linear constraint into a disjunction of linear
constraints. Then, one of the disjuncts is applied and the others are
stored, and the solving process continues; and if \unsat{} is reached,
Reluplex backtracks, removes the disjunct it has applied and applies a
different disjunct instead. The process terminates either when one of
the search paths returns \sat{} (the entire query is
\sat{}), or when they all return \unsat{} (the entire
query is \unsat{}). It is desirable to perform as few case
splits as possible, as they significantly enlarge the search space
to be explored. 

The Reluplex algorithm is formally defined as a sound and complete
calculus of derivation rules~\cite{KaBaDiJuKo21ReluplexRevised}. We omit here
the derivation rules aimed at solving the linear constraints, and
bring only the rules aimed at addressing the piecewise-linear
constraints; specifically, \relu{}
constraints~\cite{KaBaDiJuKo21ReluplexRevised}. These derivation rules are
given in Fig.~\ref{fig:abstractReluplex}, where:
\begin{inparaenum}[(i)]
\item $\allvars$ is the set of all variables in the query;
\item $\reluSet$ is the set of all \relu{} pairs; i.e., $\langle b,
  f\rangle\in\reluSet$ implies that it should hold that $f=\relu{}(b)$;
\item $\assignment$ is the current assignment, mapping variables to
  real values;
\item $l$ and $u$ map variables to their current lower and upper
  bounds, respectively; and
\item the \updateOperation($\assignment, x, v$) procedure changes
  the current assignment $\assignment$ by setting the value 
  of $x$ to $v$.
\end{inparaenum}
The $\reluCorrectB$ and $\reluCorrectF$ rules are used for correcting
an assignment in which a \relu{} constraint is currently violated, by
adjusting either the value of $b$ or $f$, respectively. The \reluSplit
rule transforms a \relu{} constraint into a disjunction, by forcing either
 $b$'s lower bound to be non-negative, or  its upper bound to
be non-positive. This forces the constraint into either its active phase
(the identity function) or its inactive phase (the zero function). In
the case when we guess that a ReLU is active, we also apply the
\addEqOperation{} operation to add the equation $f=b$, in order
to make sure the ReLU is satisfied in the active phase. The \success
rule terminates the search procedure when all variable assignments are
within their bounds (i.e., all linear constraints hold), and all
\relu{} constraints are satisfied. The rule for reaching an \unsat{}
conclusion is part of the linear constraint derivation rules which are
not depicted; see~\cite{KaBaDiJuKo21ReluplexRevised} for additional details.

\begin{figure}[t]
\begin{centering}
\scriptsize

\reluCorrectB
\drule{
\langle b, f\rangle \in \reluSet,
\ \ 
\assignment(f)\neq \relu{}(\assignment(b))
}
{
\assignment := \updateOperation(\assignment, b, \assignment(f))
}
\ \ \
\reluCorrectF
\drule{
\langle b, f\rangle \in \reluSet,
\ \ 
\assignment(f)\neq \relu{}(\assignment(b))
}
{
\assignment := \updateOperation(\assignment, f, \relu{}(\assignment(b)))
}
\medskip

\reluSplit
\drule{
\langle b, f\rangle \in \reluSet
} 
{
\begin{array}{c}
	\ub(b):=\min(\ub(b),0), \\
	\lb(f):=\max(\lb(f),0), \\
	\ub(f):=\min(\ub(f),0)
\end{array}
\qquad\qquad
\begin{array}{c}
	\lb(b):=\max(\lb(b),0), \\
	\addEqOperation(f=b)
\end{array}
}
\medskip

\success
\drule{
\forall x\in\allvars. \ 
\lb(x)\leq \assignment(x) \leq \ub(x), 
\ \ 
\forall \langle b,f \rangle \in \reluSet. \
\assignment(f) = \relu{}(\assignment(b))
} 
{
\sat{}
}
\caption{Derivation rules for the Reluplex algorithm (simplified;
  see~\cite{KaBaDiJuKo21ReluplexRevised} for more details).}
\label{fig:abstractReluplex}
\end{centering}
\end{figure}

The aforementioned rules describe a \emph{search} procedure:
the solver incrementally constructs a satisfying
assignment, and performs case splitting when needed. Another key
ingredient in modern SMT solvers is \emph{deduction} steps,
aimed at narrowing down the search space by ruling out possible
case splits. In this context, deductions are aimed at
obtaining tighter bounds for variables: i.e., finding greater
values for $l(x)$ and smaller values for $u(x)$ for each variable
$x\in\allvars$. These bounds can indeed remove case splits by fixing
activation functions into one of their phases; for example, if
$f=\relu{}(b)$ and we deduce that $b\geq 3$, we know that the \relu{}
is in its active phase, and no case split is required. We provide
additional details on some of these deduction steps in
Section~\ref{sec:optimizations}.

\section{Extending Reluplex to Support Sign Constraints}
\label{sec:supportingSigns}

In order to extend Reluplex to support \sign{}
constraints, we follow a similar approach to how \relu{}s are handled.
We encode every sign
constraint $f=\sign(b)$ as two separate
variables, $f$ and $b$. Variable $b$ represents
the input to the \sign{} function, whereas $f$
represents the \sign{}'s output. In the toy example from
Fig.~\ref{fig:toyBnn}, $b$ will represent the assignment for neuron
$v_3^1$, and $f$ will represent $v_4^1$.

Initially, a sign constraint poses no bound constraints
over $b$, i.e. $l(b)=-\infty$ and $u(b)=\infty$. Because the values of
$f$ are always $\pm{1}$, we set $l(f)=-1$ and $u(f)=1$. If,
during the search and deduction process, tighter bounds are discovered
that imply that $b\geq 0$ or $f > -1$, we say that the \sign{}
constraint has been fixed to the \emph{positive} phase; in this case,
it can be regarded as a linear constraint, namely $b\geq 0 \wedge
f=1$. Likewise, if it is discovered that $b<0$ or $f<1$, the
constraint is fixed to the \emph{negative} phase, and is regarded as
$b<0\wedge f=-1$. If neither case applies, we say that the constraint's
phase has not yet been fixed.

In each iteration of the search procedure, a violated constraint
is selected and corrected, by
altering the variable assignment.  A violated \sign{}
constraint is corrected by assigning $f$ the appropriate value:
$-1$ if the current assignment of $b$ is negative, and $1$
otherwise. Case splits (which are needed to ensure
completeness and termination) are handled similarly to the \relu{} case: we allow
the solver to assert that a sign constraint is in either the positive
or negative phase, and then backtrack and flip that assertion if the
search hits a dead-end.

More formally, we define this extension to Reluplex by modifying the
derivation rules described in Fig.~\ref{fig:abstractReluplex} as
follows. The rules for handling linear constraints and \relu{}
constraints are unchanged --- the approach is modular and extensible
in that sense, as each type of constraint is addressed separately. In
Fig.~\ref{fig:abstractSign}, we depict new derivation rules, capable of
addressing \sign{} constraints. The \signCorrectN{} and
\signCorrectP{} rules allow us to adjust the assignment of $f$ to
account for the current assignment of $b$ --- i.e., set $f$ to $-1$ if
$b$ is negative, and to $1$ otherwise. The \signSplit is used for
performing a case split on a sign constraint, introducing a
disjunction for enforcing that either $b$ is non-negative ($\lb(b)\geq 0$)
and $f=1$, or $b$ is negative ($u(b)\leq -\epsilon$; epsilon is a
small positive constant, chosen to reflect the desired precision)
and $f=-1$. Finally, the \success rule \emph{replaces} the one from
Fig.~\ref{fig:abstractReluplex}: it requires that all linear, \relu{}
and \sign{} constraints be satisfied simultaneously.

\begin{figure}[t]
  \begin{centering}
    
    \scriptsize

    \medskip
    \signCorrectN
    \drule{
      \langle b, f\rangle \in \signSet,
      \ \
      \assignment(b) < 0,
      \ \
      \assignment(f) \neq -1
    }
    {
      \assignment := \updateOperation(\assignment, f, -1)
    }
    \ \ \
    \signCorrectP
    \drule{
      \langle b, f\rangle \in \signSet,
      \ \
      \assignment(b) \geq 0,
      \ \
      \assignment(f) \neq 1
    }
    {
      \assignment := \updateOperation(\assignment, f, 1)
    }
    \medskip
    
    \signSplit
    \drule{
      \langle b, f\rangle \in \signSet 
    }
    {
 		\begin{array}{c}
                  \ub(b) := \min(\ub(b), -\epsilon), \\
                  \lb(f) := \max(\lb(f), -1), \\
                  \ub(f) := \min(\ub(f), -1)\\
 		\end{array}
 		\qquad\qquad
 		\begin{array}{c}
                  \lb(b) := \max(\lb(b), 0), \\
                  \lb(f) := \max(\lb(f), 1), \\
                  \ub(f) := \min(\ub(f), 1)\\
 		\end{array}
    }

%
%
    \medskip
    \success
    \drule{
      \forall x\in\allvars. \ 
      \lb(x) \leq \assignment(x) \leq \ub(x), \\
      \ \ 
      \forall \langle b,f \rangle \in \signSet. \
      \assignment(f)=\sign{}(\assignment(b)), \ \
      \forall \langle b,f \rangle \in \reluSet. \
      \assignment(f) = \relu{}(\assignment(b))
    }
    {
      \sat{}
    }    
      
    \caption{The extended Reluplex derivation rules, with support
      for \sign{} constraints.}
    \label{fig:abstractSign}
  \end{centering}
\end{figure}

We demonstrate this process with a simple example. Observe again the
toy example for Fig.~\ref{fig:toyBnn}, the pre-condition $P=(1\leq v_1^1\leq
  2)\wedge (-1\leq v_1^2\leq 1)$, and the post-condition
$Q=(v_5^1\leq 5)$. Our goal is to
find an assignment to the variables $\{v_1^1,v_1^2, v_2^1,
v_3^1,v_4^1,v_5^1\}$ that satisfies $P$, $Q$, and also the constraints
imposed by the BNN itself, namely the weighted sums
$v_2^1=v_1^1-v_1^2+1$, $v_3^1=0.5v_2^1$, and $v_5^1=2v_4^1$,  and the sign
constraint $v_4^1=\sign{}(v_3^1)$.

\noindent
\begin{wrapfigure}[9]{r}{5.5cm}
  \vspace{-0.8cm}
\begin{center}
\scalebox{1}{
\begin{tabular}{l|cccccc}
  variable &
  $v_1^1\ $ &
  $v_1^2\ $ &
  $v_2^1\ $ &
  $v_3^1\ $ &
  $v_4^1\ $ &
  $v_5^1\ $ 
\\
  \cline{2-7}
  assignment 1 &
  $1$ &
  $0$ &
  $2$ &
  $1$ &
  $-1$ &
  $-2$
\\
  assignment 2 &
  $1$ &
  $0$ &
  $2$ &
  $1$ &
  $\textbf{1}$ &
  $-2$              
\\
  assignment 3 &
  $1$ &
  $0$ &
  $2$ &
  $1$ &
  $1$ &
  $\textbf{2}$              
\end{tabular}
}
\caption{An iterative solution for a BNN verification query.}
\label{fig:bnnRulesApplied}
\end{center}
\end{wrapfigure}
Initially, we invoke derivation rules that address the linear
constraints (see~\cite{KaBaDiJuKo21ReluplexRevised}), and come up with an
assignment that satisfies them, depicted as assignment 1 in
Fig.~\ref{fig:bnnRulesApplied}.  However, this assignment violates
the sign constraint: $v_4^1=-1\neq\sign{}(v_3^1)=\sign{}(1)=1$. We can
thus invoke the \signCorrectP rule, which adjusts the assignment,
leading to assignment 2 in the figure.  The sign constraint is now
satisfied, but the linear constraint $v_5^1=2v_4^1$ is violated. We
thus let the solver correct the linear constraints again, this time
obtaining assignment 3 in the figure, which satisfies all
constraints. The \success rule now applies, and we return \sat{} and
the satisfying variable assignment.

The above-described calculus is sound and complete (assuming the
$\epsilon$ used in the \signSplit rule is sufficiently small): when it
answers \sat{} or \unsat{}, that statement is correct, and for any
input query there is a sequence of derivation steps that will lead to
either \sat{} or \unsat{}. The proof is quite similar to that of the
original Reluplex procedure~\cite{KaBaDiJuKo21ReluplexRevised}, and is
omitted. A naive strategy that will always lead to termination is to
apply the \signSplit rule to saturation; this effectively transforms
the problem into an (exponentially long) sequence of linear
programs. Then, each of these linear programs can be solved quickly (linear programming is known to be in \PComplexity{}).  However, this strategy is typically quite slow. In
the next section we discuss how many of these case splits can be
avoided by applying multiple optimizations.

\section{Optimizations}
\label{sec:optimizations}

\subsubsection{Weighted Sum Layer Elimination.}

\begin{figure}[htp]
  \begin{minipage}{.6\textwidth}
    \begin{center}
      \scalebox{0.8}{
        \begin{tikzpicture}[shorten >=1pt,->,draw=black!50, node distance=\layersep,font=\footnotesize]
          \def\layersep{2.2cm}

          \node[hidden neuron] (v1) at (0,-0.5) {};

          \node[hidden neuron] (v2) at (1 * \layersep, 0) {};
          \node[hidden neuron] (v3) at (1 * \layersep, -1) {};

          \node[hidden neuron] (v4) at (2 * \layersep, 0) {};
          \node[hidden neuron] (v5) at (2 * \layersep, -1) {};

          \draw[nnedge] (v1) -- node[above] {$1$} (v2);
          \draw[nnedge] (v1) -- node[below] {$-2\ \ $} (v3);

          \draw[nnedge] (v2) -- node[above] {$-1$} (v4);
          \draw[nnedge] (v2) -- node[above, pos=0.35] {$\ 3$} (v5);

          \draw[nnedge] (v3) -- node[below, pos=0.35] {$\ 2$} (v4);
          \draw[nnedge] (v3) -- node[below] {$1$} (v5);

          \draw[nnedge] ($(v1) + (-1cm, 0)$) -- (v1);

          \draw[nnedge] (v4) -- ($(v4) + (1cm, 0)$);
          \draw[nnedge] (v5) -- ($(v5) + (1cm, 0)$);

          \node[annot,above of=v2, node distance=1cm, text width=6em] (hl1) {Weighted sum
            layer \#1};
          \node[annot,above of=v4, node distance=1cm, text width=6em] (hl2) {Weighted
            sum layer \#2};

        \end{tikzpicture}
      }
    \end{center}
  \end{minipage}
  \begin{minipage}{.3\textwidth}
    \begin{center}

      \scalebox{0.8}{
        \begin{tikzpicture}[shorten >=1pt,->,draw=black!50, node distance=\layersep,font=\footnotesize]
          \def\layersep{1.5cm}

          \node[hidden neuron] (v1) at (0,-0.5) {};

          \node[hidden neuron] (v2) at (1 * \layersep, 0) {};
          \node[hidden neuron] (v3) at (1 * \layersep, -1) {};

          \draw[nnedge] (v1) -- node[above] {$-5\ \ $} (v2);
          \draw[nnedge] (v1) -- node[below] {$1$} (v3);

          \draw[nnedge] ($(v1) + (-1cm, 0)$) -- (v1);

          \draw[nnedge] (v2) -- ($(v2) + (1cm, 0)$);
          \draw[nnedge] (v3) -- ($(v3) + (1cm, 0)$);

          \node[text centered, above of=v2, node distance=1cm,text width=8em] (hl1) {Merged weighted sum
            layer};

        \end{tikzpicture}
      }

    \end{center}
  \end{minipage}
  
  \caption{On the left, a (partial) DNN with two consecutive weighted
    sum layers. On the right, an equivalent DNN with these two layers
    merged into one. 
  }
  \label{fig:mergeWsLayers}
\end{figure}
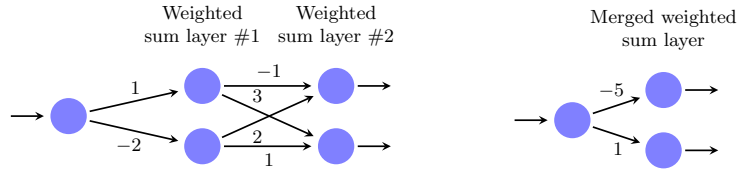

The SMT-based approach introduces a new variable for each node in a
weighted sum layer, and an equation to express that node's value
as a weighted sum of nodes from the preceding layer. In BNNs, we often
encounter consecutive weighted sum layers --- specifically because of
the binary block structure, in which a weighted sum layer is followed
by a batch normalization layer, which is also encoded as weighted sum
layer.  Thus, a straightforward way to reduce the number of variables
and equations, and hence to expedite the solution process, is to
combine two consecutive weighted sum layers into a single
layer. Specifically, the original layers can be regarded as
transforming input $x$ into $y=W_2( W_1\cdot x + B_1) + B_2$, and the
simplification as computing $y=W_3\cdot x + B_3$, where
$W_3=W_2\cdot W_1$ and $B_3=W_2\cdot B_1 + B_2$. An illustration
appears in Fig.~\ref{fig:mergeWsLayers} (for simplicity, all bias
values are assumed to be $0$).

\subsubsection{LP Relaxation.}
Given a constraint $f=\sign{}(b)$, it is 
beneficial to deduce tighter bounds on the $b$ and $f$ variables ---
especially if these tighter bounds fix the constraints into one of its
linear phases. We thus introduce a preprocessing phase, prior to the
invocation of our enhanced Reluplex procedure, in which tighter bounds
are computed by invoking a linear programming (LP) solver.

The idea, inspired by similar relaxations for \relu{}
nodes~\cite{Eh17,TjXiTe19}, is to over-approximate each constraint in
the network, including sign constraints, as a set of linear
constraints. Then, for every variable $v$ in the encoding, an LP
solver is used to compute an upper bound $u$ (by maximizing) and a
lower bound $l$ (by minimizing) for $v$.  Because the LP encoding is
an over-approximation, $v$ is indeed within the range $[l,u]$ for any
input to the network.

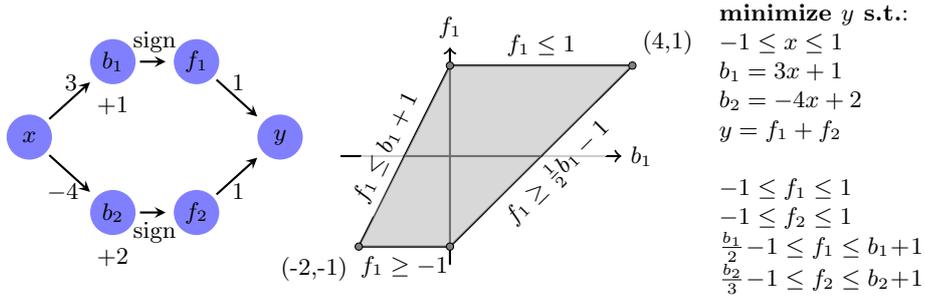
\begin{figure}[htp]
  \begin{minipage}{.35\textwidth}
    \begin{center}
      \scalebox{1.0}{
        \hspace{-0.5cm}
        \begin{tikzpicture}[shorten >=1pt,->,draw=black!50, node distance=\layersep,font=\footnotesize]
          \def\layersep{1.1cm}

          \node[hidden neuron] (v1) at (0,-1) {$x$};

          \node[hidden neuron] (v2) at (1 * \layersep, 0) {$b_1$};
          \node[hidden neuron] (v3) at (1 * \layersep, -2) {$b_2$};

          \node[below=0.05cm of v2] (b2) {$+1$};
          \node[below=0.05cm of v3] (b2) {$+2$};

          \node[hidden neuron] (v4) at (2 * \layersep, 0) {$f_1$};
          \node[hidden neuron] (v5) at (2 * \layersep, -2) {$f_2$};

          \node[hidden neuron] (v6) at (3 * \layersep, -1) {$y$};

          \draw[nnedge] (v1) -- node[above] {$3$} (v2);
          \draw[nnedge] (v1) -- node[below] {$-4\ \ $} (v3);

          \draw[nnedge] (v2) -- node[above] {\sign{}} (v4);
          \draw[nnedge] (v3) -- node[below] {\sign{}} (v5);

          \draw[nnedge] (v4) -- node[above] {$1$} (v6);
          \draw[nnedge] (v5) -- node[below] {$1$} (v6);

        \end{tikzpicture}
      }
    \end{center}
  \end{minipage}
  \begin{minipage}{.4\textwidth}
    \hspace{-1.0cm}
    \begin{tikzpicture}[scale=0.48]
      \node (tl) at (0,2.5) {}; 
      \node[label={[label distance=-0.1cm]60:(4,1)}] (tr) at (5,2.5) {}; 
      
      \node[label={[label distance=-0.1cm]210:(-2,-1)}] (bl) at (-2.5, -2.5) {}; 
      \node (br) at (0, -2.5) {}; 
     

      \draw[->,thick] (-3,0)--(4.7,0) node[right]{$b_1$};
      \draw[->,thick] (0,-3)--(0,3) node[above]{$f_1$};
      
      \draw[-,thick] (tr.center) -- node[above] {$f_1\leq 1$} (tl.center) node[left]{};
      \draw[-,thick] (bl.center) -- node[below] {$f_1\geq -1$} (br.center) node[right]{};
      
      \draw[-,thick, black] (tr.center) -- node[below, rotate=45,fill=white,
      fill opacity=0.8,text opacity=1]
      {$f_1\geq \frac{1}{2}b_1-1$} (br.center) node[right]{};
      \draw[-,thick, black] (bl.center) --
      node[above, rotate=64,fill=white,
      fill opacity=0.8,text opacity=1]
      {$f_1\leq b_1+1$} (tl.center) node[right]{};

      \fill[fill=gray!50, opacity=0.6]
      (tl.center)--(tr.center)--(br.center)--(bl.center);
            
      \draw[color=black, fill=gray] (tl) circle (.1);
      \draw[color=black, fill=gray] (tr) circle (.1);
      \draw[color=black, fill=gray] (bl) circle (.1);
      \draw[color=black, fill=gray] (br) circle (.1);

    \end{tikzpicture}

  \end{minipage}
  \begin{minipage}{.22\textwidth}
    \textbf{minimize $y$\ s.t.}:
    
    $-1\leq x \leq 1$ \\
    $b_1=3x+1$ \\
    $b_2=-4x+2$ \\
    $y=f_1 + f_2$ \\
    \\
    $-1\leq f_1 \leq 1$ \\
    $-1\leq f_2 \leq 1$ \\
    $\frac{b_1}{2} -1 \leq f_1 \leq b_1 +1$ \\
    $\frac{b_2}{3} - 1 \leq f_2 \leq b_2+1$ 
        
  \end{minipage}
  \caption{A simple BNN (left), the trapezoid relaxation of
    $f_1=\sign{}(b_1)$ (center), and its LP encoding
    (right). The trapezoid relaxation of $f_2$ is not depicted.}
  \label{fig:gurobiExample}
\end{figure}

Let $f=\sign{}(b)$, and suppose we initially know that $l\leq b\leq u$. 
The linear over-approximation that we introduce for $f$
is a trapezoid (see Fig.~\ref{fig:gurobiExample}), with the following
edges:
\begin{inparaenum}[(i)]
\item $f\leq 1$; 
\item $f\geq -1$;
\item $f\leq \frac{2}{-l}\cdot b + 1$; and
\item $f\geq \frac{2}{u}\cdot b - 1$.
\end{inparaenum}
It is straightforward to show that these four equations form the smallest
convex polytope containing the values of $f$.

We demonstrate this process on the simple BNN depicted on the left-hand side of
Fig.~\ref{fig:gurobiExample}. Suppose we know that the
input variable, $x$, is bounded in the range $-1\leq x\leq 1$, and we
wish to compute a lower bound for $y$. Simple, interval-arithmetic
based bound propagation~\cite{KaBaDiJuKo21ReluplexRevised} shows that
$b_1=3x+1$ is bounded in the range $-2\leq b_1\leq 4$, and similarly
that $b_2=-4x+2$ is in the range $-2\leq b_2\leq 6$. Because neither
$b_1$ nor $b_2$ are strictly negative or positive, we only know that
$-1\leq f_1,f_2\leq 1$, and so the best bound obtainable for $y$ is
$y\geq -2$. However, by formulating the LP relaxation of the problem
(right-hand side of Fig.~\ref{fig:gurobiExample}), we get the optimal
solution
$x=-\frac{1}{3}, b_1=0, b_2=\frac{10}{3}, f_1=-1, f_2=\frac{1}{9},
y=-\frac{8}{9}$, implying the tighter bound $y\geq -\frac{8}{9}$.

The aforementioned linear relaxation technique is effective but
expensive --- because it entails invoking the LP solver twice for each
neuron in the BNN encoding. Consequently, in our
tool, the technique is applied only once per query, as a preprocessing
step. Later, during the search procedure, we apply
a related but more lightweight technique, called \emph{symbolic bound
  tightening}~\cite{WaPeWhYaJa18}, which we enhanced to support sign
constraints.

\subsubsection{Symbolic Bound Tightening.}
In symbolic bound tightening, we compute for each neuron $v$ a
symbolic lower bound $sl(x)$ and a symbolic upper bound $su(x)$, which
are linear combinations of the input neurons. Upper and lower bounds
can then be derived from their symbolic counterparts using simple
interval arithmetic. For example, suppose the network's input nodes
are $x_1$ and $x_2$, and that for some neuron $v$ we have:
\[
  sl(v) = 5x_1 - 2x_2 + 3, \quad su(v) = 3x_1 + 4x_2 - 1
\]
and that the currently known bounds are $x_1\in[-1, 2], x_2\in [-1,1]$
and $v\in [-2,11]$. Using the symbolic bounds and the input bounds, we can
derive that the upper bound of $v$ is at most $6+4-1=9$, and that its
lower bound is at least $-5-2+3=-4$. In this case, the upper bound we
have discovered for $v$ is tighter than the previous one, and so we
can update $v$'s range to be $[-2,9]$.

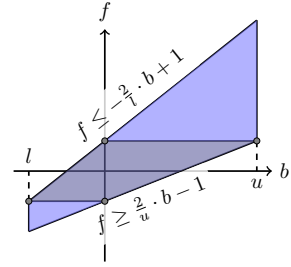
\begin{wrapfigure}[13]{r}{4.5cm}
  \vspace{-1.0cm}
  \begin{center}
    \scalebox{0.8} {
      \begin{tikzpicture}[scale=0.50]
        \node (tl) at (0,1) {}; 
        \node (tr) at (5,1) {}; 
        
        \node (bl) at (-2.5, -1) {}; 
        \node (br) at (0, -1) {}; 

        \node (b) at (-2.5, -2) {};
        \node (t) at (5, 5) {};
        

        \draw[->,thick] (-3,0)--(5.5,0) node[right]{$b$};
        \draw[->,thick] (0,-3)--(0,4.7) node[above]{$f$};
        
        \draw[-,thick] (tr.center) -- node[above] {} (tl.center) node[left]{};
        \draw[-,thick] (bl.center) -- node[below] {} (br.center) node[right]{};

        \draw[-,thick, black] (tr.center) -- node[below, rotate=21.8,fill=white,
        fill opacity=0.8,text opacity=1]
        {$f\geq \frac{2}{u}\cdot b-1$} (b.center) node[right]{};
        
        \draw[-,thick, black] (bl.center) --
        node[above, rotate=38.5,fill=white,
        fill opacity=0.8,text opacity=1]
        {$f\leq -\frac{2}{l}\cdot b+1$} (t.center) node[right]{};

        \draw[-,thick] (t.center) -- (tr.center);
        \draw[-,thick,dashed] (tr.center) -- (5,0) node[below]{$u$};

        \draw[-,thick] (b.center) -- (bl.center);
        \draw[-,thick,dashed] (bl.center) -- (-2.5,0) node[above]{$l$};

        \fill[fill=gray!75!blue, opacity=0.6]
        (tl.center)--(tr.center)--(br.center)--(bl.center);

        \fill[fill=blue!50, opacity=0.6]
        (t.center)--(tr.center)--(tl.center);

        \fill[fill=blue!50, opacity=0.6]
        (b.center)--(br.center)--(bl.center);

        \draw[color=black, fill=gray] (tl) circle (.1);
        \draw[color=black, fill=gray] (tr) circle (.1);
        \draw[color=black, fill=gray] (bl) circle (.1);
        \draw[color=black, fill=gray] (br) circle (.1);
      \end{tikzpicture}
    }
  \end{center}
  \caption{Symbolic bounds for f=\sign{}(b).}
  \label{fig:sbtTrapezoid}  
\end{wrapfigure}

The symbolic bound expressions are propagated layer by
layer~\cite{WaPeWhYaJa18}. Propagation through weighted sum layers is
straightforward: the symbolic bounds are simply multiplied by the
respective edge weights and summed up. Efficient approaches for
propagations through ReLU layers have also been
proposed~\cite{KuKaGoJuBaKo18b}. Our contribution here is an extension
of these techniques for propagating symbolic bounds also through
\sign{} layers. The approach again uses a trapezoid, although a more
coarse one --- so that we can approximate each neuron from above and
below using a single linear expression. More specifically, for
$f=\sign{}(b)$ with $b\in[l,u]$ and previously-computed symbolic
bounds $su(b)$ and $sl(b)$, the symbolic bounds for $f$ are given by:
\[
  sl(f) = \frac{2}{u}\cdot sl(b) - 1, \quad  su(f) = -\frac{2}{l}\cdot su(b) + 1
\]
An illustration appears in Fig.~\ref{fig:sbtTrapezoid}. The blue
trapezoid is the relaxation we use for the symbolic bound computation,
whereas the gray trapezoid is the one used for the LP relaxation
discussed previously. The blue trapezoid is larger, and hence leads to
looser bounds than the gray trapezoid; but it is computationally
cheaper to compute and use, and our evaluation demonstrates its
usefulness.

\subsubsection{Polarity-based Splitting.}
The Marabou framework supports a parallelized solving mode, using the
Split-and-Conquer (\snc{}) algorithm~\cite{WuOzZeIrJuGoFoKaPaBa20}.
At a high level, \snc{} partitions a verification query $\phi$ into a
set of sub-queries $\Phi:=\{\phi_1,...\phi_n\}$, such that $\phi$ and
$\bigvee_{\phi'\in\Phi}\phi'$ are equi-satisfiable, and handles each
sub-query independently. Each sub-query is solved with a timeout
value; and if that value is reached, the sub-query is again split into
additional sub-queries, and each is solved with a greater timeout
value. The process repeats until one of the sub-queries is determined
to be \sat{}, or until all sub-queries are proven \unsat{}.

One Marabou strategy for creating sub-queries is by splitting
the ranges of input neurons. For example, if in query $\phi$ an input
neuron $x$ is bounded in the range $x\in [0,4]$ and $\phi$ times out,
it might be split into $\phi_1$ and $\phi_2$ such that $x\in [0,2]$ in
$\phi_1$ and $x\in [2,4]$ in $\phi_2$. This strategy is effective when
the neural network being verified has only a few input neurons.

Another way to create sub-queries is to perform case-splits on
piecewise-linear constraints --- sign constraints, in our case. For
instance, given a verification query
$\phi := \phi' \land f=\sign{}(b)$, we can partition it into
$\phi^{-} := \phi' \land b < 0 \land f = -1$ and
$\phi^{+} := \phi' \land b \geq 0 \land f = 1$. Note that $\phi$ and
$\phi^{+} \lor \phi^{-}$ are equi-satisfiable.

The heuristics for picking which sign constraint to split on have a
significant impact on the difficulty of the resulting
sub-problems~\cite{WuOzZeIrJuGoFoKaPaBa20}.  Specifically, it is
desirable that the sub-queries be \emph{easier} than the original query, and
also that they be \emph{balanced} in terms of runtime --- i.e., we
wish to avoid the case where $\phi_1$ is very easy and $\phi_2$ is
very hard, as that makes poor use of parallel computing resources. To create
easier sub-problems, we propose to split on sign constraints that
occur in the earlier layers of the BNN, as that leads to efficient
bound propagation when combined with our symbolic bound tightening
mechanism.  To create balanced sub-problems, we
use a metric called \emph{polarity}, which was proposed in
\cite{WuOzZeIrJuGoFoKaPaBa20} for \relu{}s and is extended here to
support sign constraints.

\begin{definition}
  \label{def:pol}
Given a sign constraint $f=sign{}(b)$, and the bounds $l \leq b \leq u$,
where $l < 0$, and $u > 0$, the polarity of the sign constraint is
defined as $p = \frac{u + l}{u - l}$.
\end{definition}
\noindent

Intuitively, the closer the polarity is to 0, the more balanced the
resulting queries will be if we perform a case-split on this
constraint. For example, if $\phi = \phi'\wedge -10\leq b\leq 10$ and
we create $\phi_1=\phi'\wedge -10\leq b<0$,
$\phi_2=\phi'\wedge 0\leq b\leq 10$, then queries $\phi_1$ and
$\phi_2$ are roughly balanced. However, if initially
$-10\leq b \leq 1$, we obtain $\phi_1=\phi'\wedge -10\leq b< 0$ and
$\phi_2=\phi'\wedge 0\leq b\leq 1$. In this case, $\phi_2$ might prove
significantly easier than $\phi_1$ because the smaller range of $b$
in $\phi_2$ could lead to very effective bound tightening.
Consequently, we use a heuristic that picks the sign
constraint with the smallest polarity among the
first $k$ candidates (in topological order), where $k$
is a configurable parameter. In our experiments, we empirically
selected $k=5$. 

\section{Implementation}
\label{sec:tool}

We implemented our approach as an extension to
Marabou~\cite{KaHuIbJuLaLiShThWuZeDiKoBa19Marabou}, which is an
open-source, freely available SMT-based DNN verification
framework~\cite{MarabouGitRepo}. Marabou implements the Reluplex algorithm, but with
multiple extensions and optimizations --- e.g., support for additional
activation functions, deduction methods, and
parallelization~\cite{WuOzZeIrJuGoFoKaPaBa20}. It has been used for a
variety of verification tasks, such as network
simplification~\cite{GoFeMaBaKa20} and
optimization~\cite{StWuZeJuKaBaKo20}, verification of video streaming
protocols~\cite{KaBaKaSc19}, DNN modification~\cite{GoAdKeKa20},
adversarial robustness
evaluation~\cite{KaBaDiJuKo17Fvav,CaKaBaDi17,GoKaPaBa18} verification
of recurrent networks~\cite{JaBaKa20}, and others.  However, to date
Marabou could not support \sign{} constraints, and thus, could not be
used to verify BNNs. Below we describe our main contributions to the
code base.  Our complete code is available as an artifact accompanying
this paper~\cite{ArtifactGitRepo}, and has also been  merged into the main Marabou
repository~\cite{MarabouGitRepo}.

\subsubsection{Basic Support for Sign Constraints (\emph{SignConstraint.cpp}).}
During execution, Marabou maintains a set of piecewise-linear
constraints that are part of the query being solved. To support
various activation functions, these constraints are represented using
classes that inherit from the abstract
\emph{PiecewiseLinearConstraint} class. Here, we added a new
sub-class, \emph{SignConstraint}, that inherits from
\emph{PiecewiseLinearConstraint}. The methods of this class check
whether the piecewise-linear sign constraint is satisfied, and in case
it is not --- which possible changes to the current assignment could
fix the violation. This class' methods also extend Marabou's deduction
mechanism for bound tightening.

\sloppy
\subsubsection{Input Interfaces for Sign Constraints (\emph{MarabouNetworkTF.py}).}
Marabou supports various input interfaces, most notable of which is
the TensorFlow interface, which automatically translates a DNN stored
in TensorFlow \textit{protobuf} or \textit{savedModel} formats into a
Marabou query. As part of our extensions, we enhanced this interface so
that it can properly handle BNNs and \sign{} constraints.
Additionally, users can create queries using Marabou's native C++
interface, by instantiating the \emph{SignConstraint} class discussed
previously.

\subsubsection{Network-Level Reasoner (\emph{NetworkLevelReasoner.cpp,
  Layer.cpp, LPFormulator.cpp}).}

The \emph{Network-Level Reasoner} (\emph{NLR}) is the part of Marabou
that is aware of the topology of the neural network being verified, as
opposed to just the individual constraints that comprise it.  We
extended Marabou's NLR to support \sign{} constraints and implement
the optimizations discussed in
Section~\ref{sec:optimizations}. Specifically, one extension that we
added allows this class to identify consecutive weighted sum layers
and merge them.  Another extension creates a linear over-approximation
of the network, including the trapezoid-shaped over-approximation of
each sign constraint.  As part of the symbolic bound propagation
process, the NLR traverses the network, layer by layer, each time
computing the symbolic bound expressions for each neuron in the
current layer.

\subsubsection{Polarity-Based Splitting (\emph{DnCManager.cpp}).}
We extended the methods of this class, which is part of Marabou's
\snc{} mechanism, to compute the polarity value of each \sign{}
constraint (see Definition \ref{def:pol}), based on the current bounds.

\section{Evaluation}
\label{sec:evaluation}

All the benchmarks described in this section are included in our
artifact, and are publicly available online~\cite{ArtifactGitRepo}.

\subsubsection{Strictly Binarized Networks.}
We began by training a strictly binarized network over the MNIST digit
recognition dataset.\footnote{\url{http://yann.lecun.com/exdb/mnist/}} This
dataset includes 70,000 images of handwritten digits, each given as a
$28\times28$ pixeled image, with normalized brightness values ranging from 0 to 1. 
The network that we trained has an input layer of size 784, followed by six 
binary blocks (four blocks of size $50$, two blocks of size $10$), and a final
output layer with 10 neurons.  Note that in the first block we omitted
the \sign{} layer in order to improve the network's
accuracy.\footnote{This is standard practice; see
  \url{https://docs.larq.dev/larq/guides/bnn-architecture/}} The model
was trained for 300 epochs using the \emph{Larq}
library~\cite{geiger2020larq} and the \emph{Adam} optimizer~\cite{KiBa2014}, 
achieving 90\% accuracy.

\begin{wrapfigure}[9]{r}{5.8cm}
  \vspace{-1.3cm}
  \begin{center}
    \includegraphics[width=28mm]{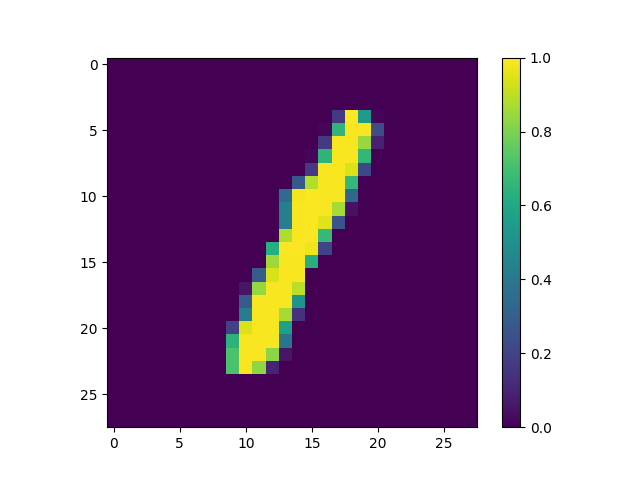}
    \includegraphics[width=28mm]{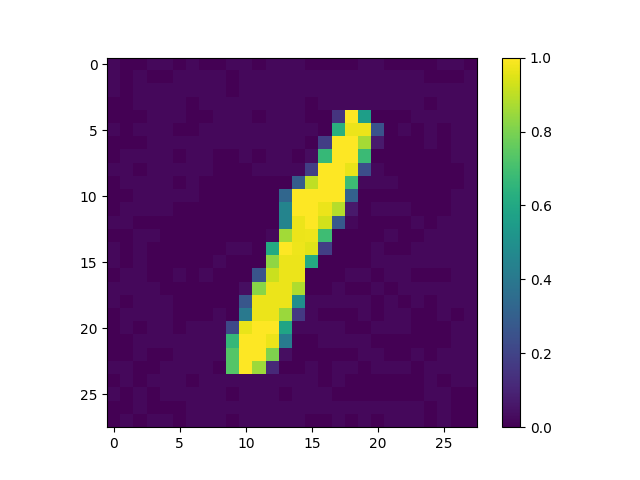}
   
  \end{center}
  \caption{An adversarial example for the MNIST network.}
  \label{fig:mnistAdvExamples}
\end{wrapfigure}
After training, we used Larq's export mechanism to save the trained
network in a TensorFlow format, and then used our newly added Marabou
interface to load it. For our verification queries, we first chose 500
samples from the test set which were classified correctly by the
network. Then, we used these samples to formulate \emph{adversarial
  robustness} queries~\cite{SzZaSuBrErGoFe13, KaBaDiJuKo21ReluplexRevised}:
queries that ask Marabou to find a slightly perturbed input which is
misclassified by the network, i.e. is assigned a different label than
the original. We formulated 500 queries, constructed from 50 queries
for each of ten possible perturbation values
$\delta \in \{0.1, 0.15, 0.2, 0.3, 0.5, 1, 3, 5, 10, 15\}$ in
$L_\infty$ norm, one query per input sample. An \unsat{} answer from
Marabou indicates that no adversarial perturbation exists (for the
specified $\delta$), whereas a \sat{} answer includes, as the
counterexample, an actual perturbation that leads to
misclassification. Such adversarial robustness queries are the most
widespread verification benchmarks in the literature
(e.g.,~\cite{KaBaDiJuKo21ReluplexRevised, GeMiDrTsChVe18, WaPeWhYaJa18,HuKwWaWu17}).
An example appears in Fig.~\ref{fig:mnistAdvExamples}: the image on
the left is the original, correctly classified as 1, and the image on
the right is the perturbed image discovered by Marabou, 
misclassified as 3.

Through our experiments we set out to evaluate our tool's performance,
and also measured the contribution of each of the features that we
introduced:
\begin{inparaenum}[(i)]
\item weighted sum (ws) layer elimination;
\item LP relaxation;
\item symbolic bound tightening (sbt); and
\item polarity-based splitting.
\end{inparaenum}
We thus defined five configurations of the tool: the \emph{all}
category, in which all four features are enabled, and four
\emph{all-X} configurations for \emph{X}$\in\{$\emph{ws, lp, sbt,
  polarity}$\}$, indicating that feature X is turned off and the other
features are enabled.  All five configurations utilized Marabou's
parallelization features, except for \emph{all-polarity} --- where
instead of polarity-based splitting we used Marabou's default splitting 
strategy, which splits the input domain in half in each step.

\begin{figure}[htp]
  \begin{center}
   \includegraphics[width=7.5cm]{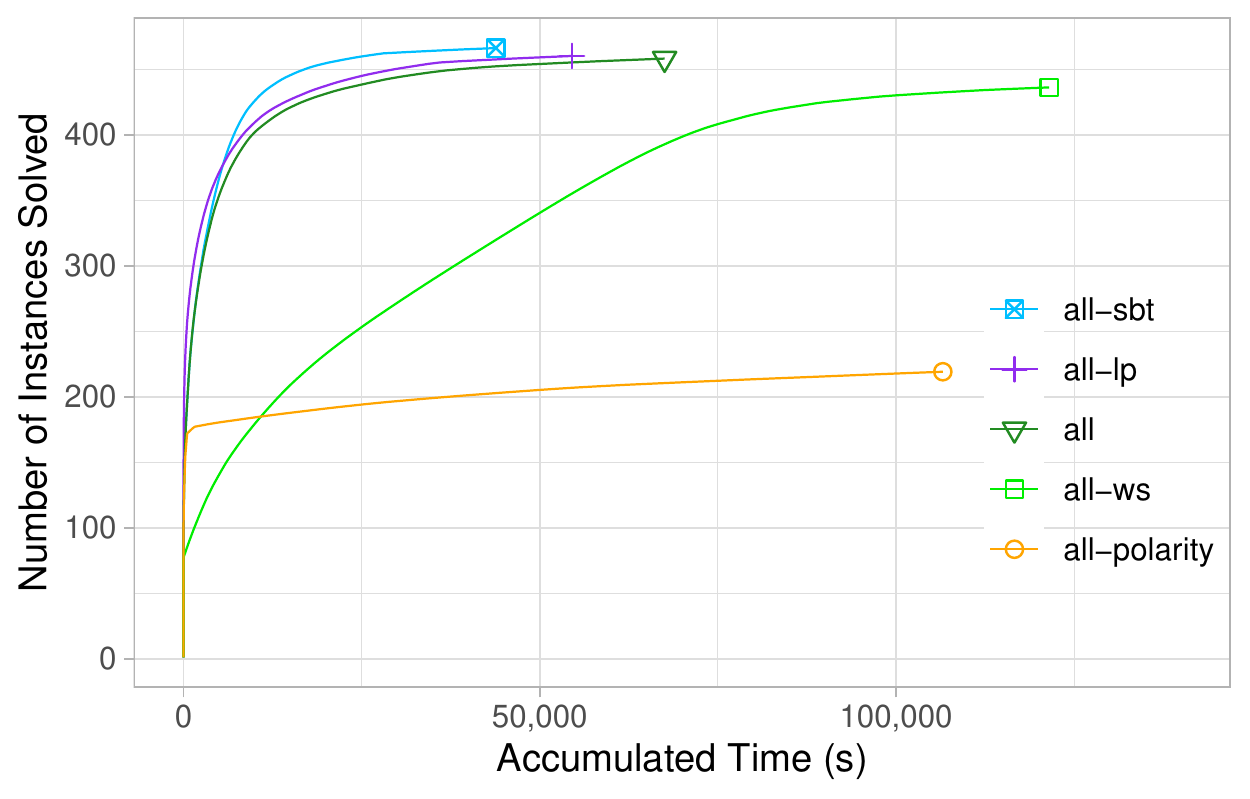}
  \end{center}
  \caption{Running the five configurations of Marabou on the MNIST BNN.}
  \label{fig:compareConfigurations}
\end{figure}

\begin{figure}[htp]
     \begin{minipage}{0.5\textwidth}
        \centering
        \includegraphics[width=5.25cm]{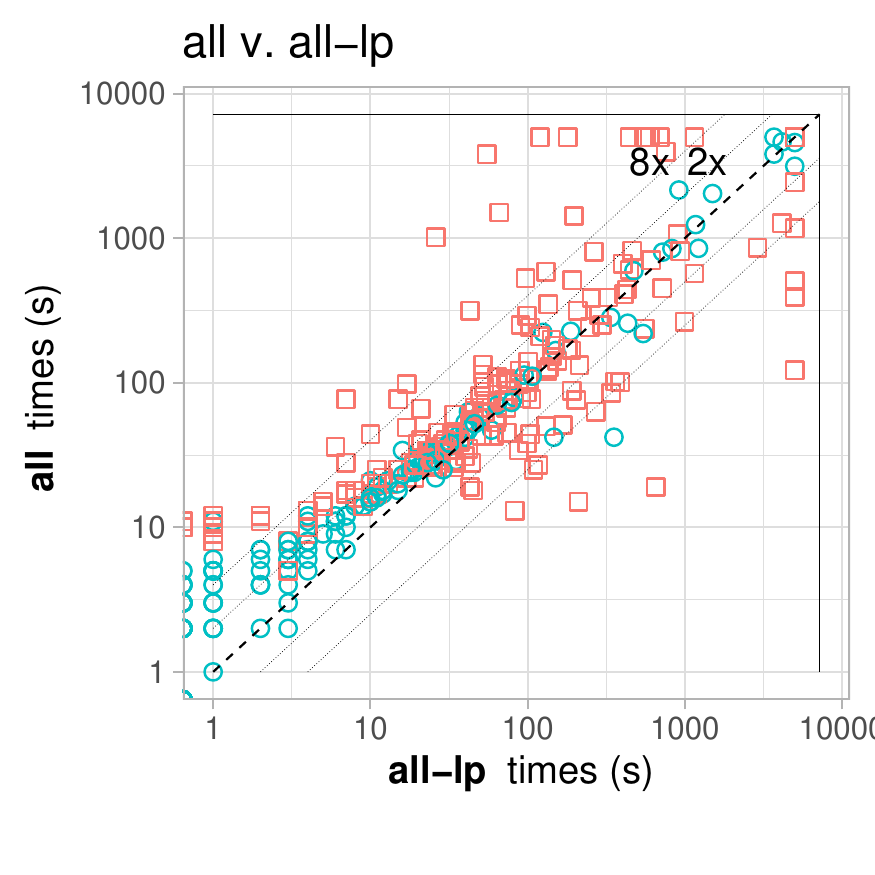}
    \end{minipage}%
    \begin{minipage}{0.5\textwidth}
        \centering
        \includegraphics[width=6.5cm]{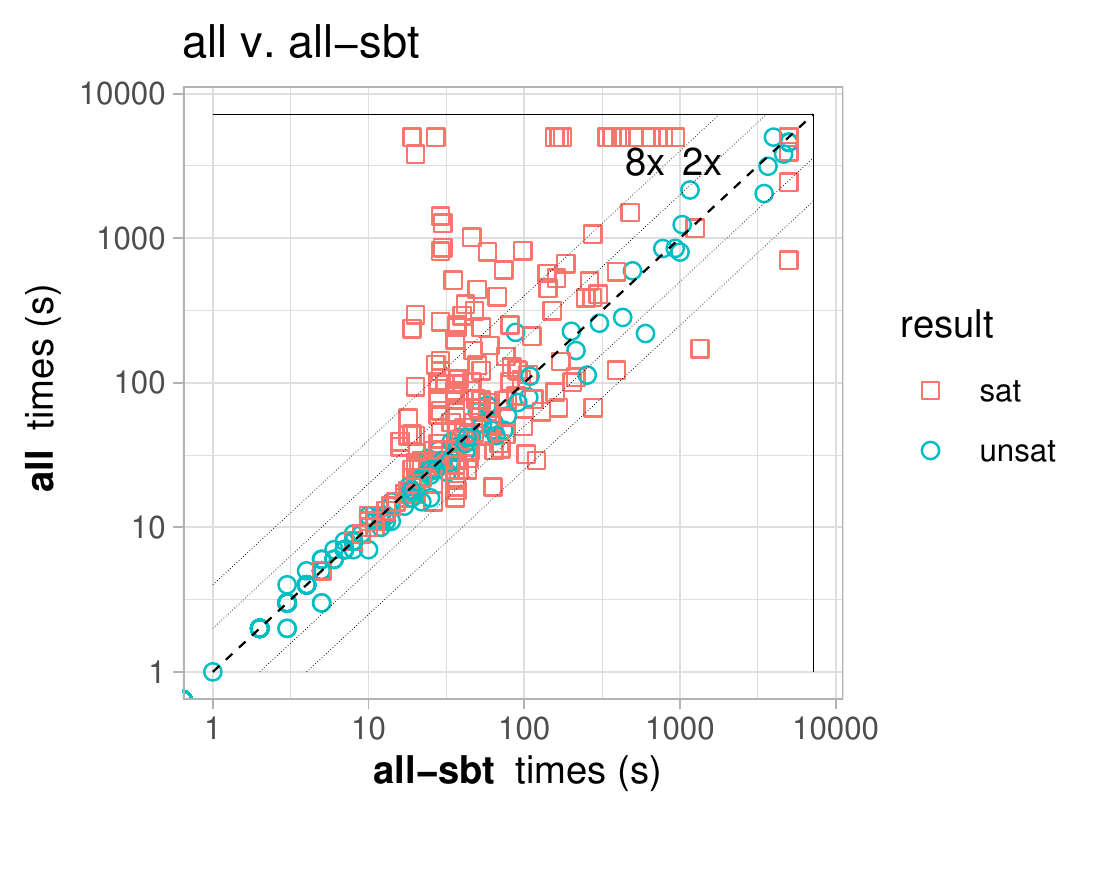}
    \end{minipage}
  \caption{Evaluating the LP relaxation and symbolic bound tightening features.}
  \label{fig:compareSbtAndLP}
\end{figure}

Fig.~\ref{fig:compareConfigurations} depicts Marabou's results using
each of the five configurations. Each experiment was run on an Intel
Xeon E5-2637 v4 CPUs machine, running Ubuntu 16.04 and using eight
cores, with a wall-clock timeout of 5,000 seconds.  Most notably, the
results show the usefulness of polarity-based splitting when compared
to Marabou's default splitting strategy: whereas the
\emph{all-polarity} configuration only solved 218 instances, the
\emph{all} configuration solved 458. It also shows that the weighted
sum layer elimination feature significantly improves performance, from
436 solved instances in \emph{all-ws} to 458 solved instances in
\emph{all}, and with significantly faster solving speed. With the
remaining two features, namely LP relaxations and symbolic bound
tightening, the results are less clear: although the \emph{all-lp} and
\emph{all-sbt} configurations both slightly outperform the \emph{all}
configuration, indicating that these two features slowed down the
solver, we observe that for many instances they do lead to an
improvement; see Fig.~\ref{fig:compareSbtAndLP}. Specifically, on
\unsat{} instances, the \emph{all} configuration was able to solve one
more benchmark than either \emph{all-lp} or \emph{all-sbt}; and it
strictly outperformed \emph{all-lp} on 13\% of the instances, and
\emph{all-sbt} on 21\% of the instances. Gaining better insights into
the causes for these differences is a work in progress.

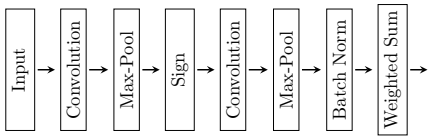
\begin{wrapfigure}[7]{r}{6.8cm}
  \vspace{-0.5cm}
  \begin{center}
    \scalebox{0.7}{
    \begin{tikzpicture}
      [block/.style={
          draw,
          shape=rectangle,
          minimum width=2.3cm,
          align=center,
          rotate=90,
        }]
        
        \node[block] (A) at (0,0) {Input};
        \node[block] (B) at (1,0) {Convolution};
        \node[block] (C) at (2,0) {Max-Pool};
        \node[block] (D) at (3,0) {Sign};
        \node[block] (E) at (4,0) {Convolution};
        \node[block] (F) at (5,0) {Max-Pool};
        \node[block] (G) at (6,0) {Batch Norm};
        \node[block] (H) at (7,0) {Weighted Sum};
        \node[block,color=white] (Fake) at (8,0) {A};

        \draw[nnedge] (A.south) -- (B.north);
        \draw[nnedge] (B.south) -- (C.north);
        \draw[nnedge] (C.south) -- (D.north);
        \draw[nnedge] (D.south) -- (E.north);
        \draw[nnedge] (E.south) -- (F.north);
        \draw[nnedge] (F.south) -- (G.north);
        \draw[nnedge] (G.south) -- (H.north);
        \draw[nnedge] (H.south) -- (Fake.north); 

      \end{tikzpicture}
      }
    \caption{The XNOR-Net architecture of our network.}
    \label{fig:xnorNet}
  \end{center}
\end{wrapfigure}

\subsubsection{XNOR-Net.}
XNOR-Net~\cite{rastegari2016xnor} is a BNN architecture for image
recognition networks.  XNOR-Nets consist of a series of \emph{binary
  convolution} blocks, each containing a \sign{} layer, 
 a convolution layer, and a max-pooling layer (here, 
we regard convolution layers as a specific case of weighted sum
layers). We constructed such a network with two binary convolution
blocks: the first block has three layers, including a convolution
layer with three filters, and the second block has four layers,
including a convolution layer with two filters.  The two binary
convolution blocks are followed by a batch normalization layer and a
fully-connected weighted sum layer (10 neurons) for the network's
output, as depicted in Fig.~\ref{fig:xnorNet}.  Our network was
trained on the Fashion-MNIST dataset, which includes 70,000 images
from ten different clothing categories~\cite{xiao2017fashion}, each
given as a $28\times28$ pixeled image. The model was trained for 30
epochs, and achieved a modest accuracy of 70.97\%.

\begin{wrapfigure}[9]{r}{5.8cm}
  \vspace{-0.6cm}
	\centering
	\includegraphics[width=28mm]{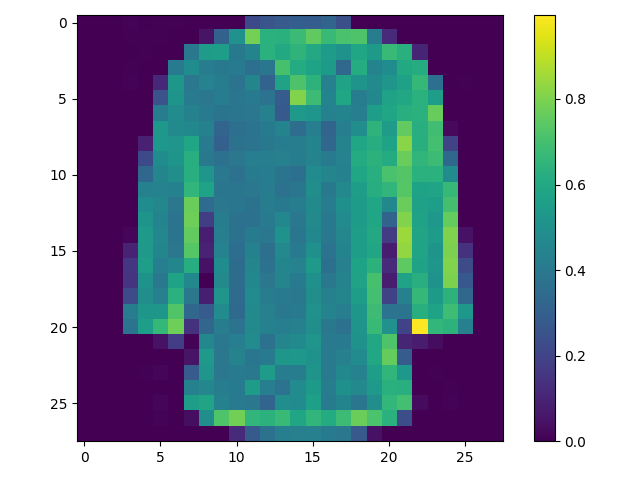}
	\includegraphics[width=28mm]{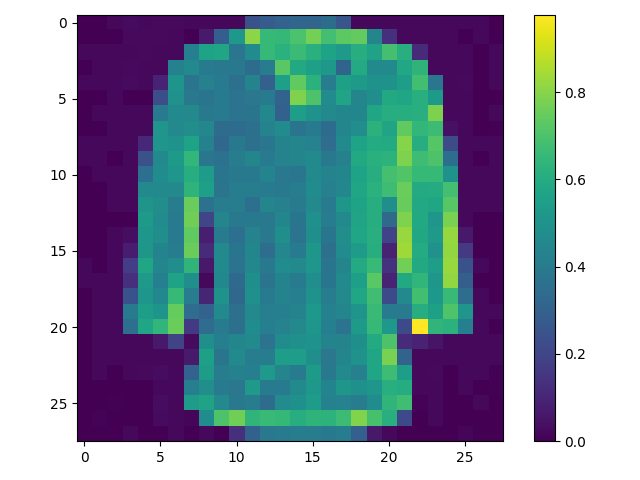}
	
	\caption{An original image (left) and its perturbed,
          misclassified image (right).}
	\label{fig:xnorAdvExample}
\end{wrapfigure}
For our verification queries, we chose 300 correctly classified
samples from the test set, and used them to formulate adversarial
robustness queries. Each query was formulated using one sample and a
perturbation value $\delta \in \{0.05, 0.1, 0.15, 0.2, 0.25, 0.3\}$ in
$L_\infty$ norm.  Fig.~\ref{fig:xnorAdvExample} depicts the
adversarial image that Marabou produced for one of these queries.  The 
image on the left is a correctly classified image of a shirt, and the
image on the right is the perturbed image, now misclassified as a
coat.

Based on the results from the previous set of experiments, we used
Marabou with weighted sum layer elimination and polarity-based
splitting turned on, but with symbolic bound tightening and LP
relaxation turned off. Each experiment ran on an Intel Xeon E5-2637 v4 machine,
using eight cores and a wall-clock timeout of 7,200 seconds. The
results are depicted in Table~\ref{table:xnorNetResults}.  The results
demonstrate that \unsat{} queries tended to be solved significantly
faster than \sat{} ones, indicating that Marabou's search
procedure for these cases needs further optimization. Overall, Marabou
was able to solve 203 out of 300 queries.  To the best of our
knowledge, this is the first effort to formally verify
an XNOR-Net. We note that these results demonstrate the usefulness
of an SMT-based approach for BNN verification, as it allows the verification of
DNNs with multiple types of activation functions, such as a
combination of \sign{} and max-pooling.

\begin{table}[t]
  \centering
  \caption{
    Marabou's performance on the XNOR-Net queries.
  }
  \scalebox{0.8}{
    \begin{tabular}[htp]{lc|crcrcrcrcr}
      \toprule
      \multirow{3}{*}{\vspace*{8pt}\hspace*{9pt}$\delta$}
      &&& 
          \multicolumn{3}{c}{\sat}
      &&  
         \multicolumn{3}{c}{\unsat}
      &&
      \\
      
       \cline{4-6}
      \cline{8-10}
      &&&  
          \# Solved && Avg. Time (s)
      &&  
         \# Solved && Avg. Time (s)
        &&
           \# Timeouts
      \\
      
      \midrule
      0.05
      &&&
         15 && 909.13
      &&
         23 && 4.96 
        &&
           12
      \\
      
      0.1
      &&&
          15 && 1,627.67
      &&
         20 && 12.15
        &&
           15
      \\
      
      0.15
      &&&
          9 && 1,113.33
      &&
         29 && 5
        &&
           12
      \\
      
      0.2
      &&&
          10 && 1,387.7  
      &&
         24 && 4.96
        &&
           16
      \\
      
      0.25
      &&&
          9 && 1,426
      &&
         22 && 4.91
        &&
           19
      \\
      
      0.3
      &&&
          7 && 1,550.86
      &&
         20 && 26.75
        &&
           23
      \\
      
      \midrule
      Total
      &&&
          65 && 1,317.52
      &&
         138 && 9.16
        &&
           97
      \\
			
      \bottomrule
    \end{tabular}%
  } 
  \label{table:xnorNetResults}
\end{table}%

\section{Related Work}
\label{sec:relatedWork}

DNNs have become pervasive in recent years, and the discovery of
various faults and errors has given rise to multiple approaches for
verifying them. These include various SMT-based approaches
(e.g.,~\cite{HuKwWaWu17, KaBaDiJuKo21ReluplexRevised,
  KaHuIbJuLaLiShThWuZeDiKoBa19Marabou, KuKaGoJuBaKo18}), approaches
based on LP and MILP solvers (e.g.,~\cite{LoMa17,Eh17,TjXiTe19,
  BuTuToKoMu18}), approaches based on symbolic interval propagation or
abstract interpretation (e.g.,~\cite{WaPeWhYaJa18, GeMiDrTsChVe18,
  WeZhChSoHsBoDhDa18, TrBkJo20}), abstraction-refinement
(e.g.,~\cite{ElGoKa20, AsHaKrMu20}), and many others.  Most of these
lines of work have focused on non-quantized DNNs. Verification of
quantized DNNs is \PSPACEComplexity{}-hard~\cite{HeKeZiScalable20}, and requires
different tools than the ones used for their non-quantized
counterparts~\cite{GiMiHeThMa20}.  Our technique extends an existing
line of SMT-based verifiers to support also the \sign{} activation
functions needed for verifying BNNs; and these new activations can be
combined with various other layers.

Work to date on the verification of BNNs has
relied exclusively on reducing the problem to Boolean satisfiability,
and has thus been limited to the strictly binarized
case~\cite{NaZhGuWa19, narodytska2017verifying, jia2020efficient, ChNuHuRu17}. 
Our approach, in contrast, can be applied to binarized neural networks that include
activation functions beyond the \sign{} function, as we have
demonstrated by verifying an XNOR-Net. Comparing the performance of
Marabou and the SAT-based approaches is left for future work.
  
\section{Conclusion}
\label{sec:conlcusion}

BNNs are a promising avenue for leveraging deep learning in devices
with limited resources. However, it is highly desirable to verify
their correctness prior to deployment. Here, we propose an
 SMT-based verification approach that enables the
verification of BNNs. This approach, which we
have implemented as part of the Marabou framework~\cite{MarabouGitRepo}, seamlessly
integrates with the other components of the SMT solver in a
modular way. Using Marabou, we have verified, for the first time, a
network that uses both binarized and non-binarized layers. In the
future, we plan to improve the scalability of our approach, by
enhancing it with stronger bound deduction capabilities, based on
abstract interpretation~\cite{GeMiDrTsChVe18}.

\subsubsection{Acknowledgements.}
We thank Nina Narodytska, Kyle Julian, Kai Jia, Leon Overweel and the Plumerai 
research team for their contributions to this project.  
The project was partially supported by the Israel Science
Foundation (grant number 683/18), the Binational Science Foundation
(grant number 2017662), the National Science Foundation (grant
number 1814369), and the Center for Interdisciplinary Data Science Research at The Hebrew University of Jerusalem.

{
\newpage
\bibliographystyle{abbrv}
\bibliography{bnns}
}

\end{document}